\documentclass{article}

     \PassOptionsToPackage{numbers, compress}{natbib}

     \usepackage[preprint]{neurips_2022}

\usepackage[utf8]{inputenc} 
\usepackage[T1]{fontenc}    
\usepackage{hyperref}       
\usepackage{url}            
\usepackage{booktabs}       
\usepackage{amsfonts}       
\usepackage{nicefrac}       
\usepackage{microtype}      
\usepackage{xcolor}         

\usepackage{graphicx}  
\usepackage{wrapfig}  
\usepackage{amsmath}  
\usepackage{subcaption}

\usepackage{pifont}
\newcommand{\cmark}{\ding{51}}%
\newcommand{\xmark}{\ding{55}}%

\title{Guidelines for the Regularization of Gammas in Batch Normalization for Deep Residual Networks}

\author{%
    Bum Jun Kim \\
    POSTECH \\
    \texttt{kmbmjn@postech.edu} \\
    \And
    Hyeyeon Choi \\
    POSTECH \\
    \texttt{hyeyeon@postech.edu} \\
    \AND
    Hyeonah Jang \\
    POSTECH \\
    \texttt{hajang@postech.edu} \\
    \And
    Dong Gu Lee \\
    POSTECH \\
    \texttt{dgleee@postech.edu} \\
    \And
    Wonseok Jeong \\
    POSTECH \\
    \texttt{wonseok.jeong@postech.edu} \\
    \And
    Sang Woo Kim \\
    POSTECH \\
    \texttt{swkim@postech.edu} \\
}

\begin{document}

\maketitle

\begin{abstract}
	$L_2$ regularization for weights in neural networks is widely used as a standard training trick. However, $L_2$ regularization for $\gamma$, a trainable parameter of batch normalization, remains an undiscussed mystery and is applied in different ways depending on the library and practitioner. In this paper, we study whether $L_2$ regularization for $\gamma$ is valid. To explore this issue, we consider two approaches: 1) variance control to make the residual network behave like identity mapping and 2) stable optimization through the improvement of effective learning rate. Through two analyses, we specify the desirable and undesirable $\gamma$ to apply $L_2$ regularization and propose four guidelines for managing them. In several experiments, we observed the increase and decrease in performance caused by applying $L_2$ regularization to $\gamma$ of four categories, which is consistent with our four guidelines. Our proposed guidelines were validated through various tasks and architectures, including variants of residual networks and transformers.
\end{abstract}

\section{Introduction}
\label{sec:introduction}
Deep neural networks have exhibited remarkable performance in various fields. Previously, large neural networks with deep and wide architectures were considered difficult to train. However, various regularization techniques such as $L_2$ regularization \cite{giclr/ZhangWXG19}, batch normalization (BN) \cite{gicml/IoffeS15}, and residual learning \cite{gcvpr/HeZRS16} have made it possible to train large neural networks, leading to successful performance.

Since the classic machine learning era, $L_2$ regularization has been applied as a restriction on the weights $W$ of neural networks and has not been applied to bias $b$. However, trainable parameters in modern neural networks are not limited to weights $W$ and bias $b$. BN outputs $\gamma_i \hat{x_i} + \beta_i$ from normalized feature $\hat{x}$, where $\gamma$ and $\beta$ are also trainable parameters. Because $\beta$ plays a similar role to bias $b$, we ignore $L_2$ regularization of $\beta$. However, $\gamma$ controls the scale of the feature map and can be viewed as a special case of weights (Section \ref{sec:effectivelearningrateanalysis}).
Despite the similar roles of $\gamma$ and weights, $L_2$ regularization of the $\gamma$ parameters of BN remains an undiscussed mystery. Moreover, each deep learning library and practice provides a different method:
\paragraph{TensorFlow and Keras} To implement $L_2$ regularization, a regularizer option needs to be set for each layer. For example, \texttt{kernel\_regularizer} can be applied to the convolution and fully connected layer to implement $L_2$ regularization on weight $W$. $L_2$ regularization of $\gamma$ is functionally supported through \texttt{gamma\_regularizer}, which may be set to the BN layer. In practice, however, \texttt{gamma\_regularizer} is rarely used. To our knowledge, in the TensorFlow and Keras official tutorials and code, practical use of \texttt{gamma\_regularizer} does not exist. In other words, $L_2$ regularization in TensorFlow and Keras generally means using \texttt{kernel\_regularizer} on weights, and $L_2$ regularization has not been applied to $\gamma$.
\paragraph{PyTorch} $L_2$ regularization is implemented by applying a weight decay at the optimizer level. However, the weight decay of PyTorch is applied to all trainable parameters, including all $W$, $b$, $\gamma$, and $\beta$. Although the validity of weight decay for $b$ and $\beta$ is also arguable, we focus on whether PyTorch's practice of applying weight decay to $\gamma$ is valid.
\paragraph{Practice and empirical observation} In several work \cite{cvpr/HeZ0ZXL19,corr/abs-1807-11205,gcorr/GoyalDGNWKTJH17}, $L_2$ regularization was not applied $\gamma$ and $\beta$. However, \cite{ijcv/WuH20,iclr/YanWZ0W020} used $L_2$ regularization on $\gamma$ and $\beta$, while \cite{ijcv/WuH20} turned off it during fine-tuning. Reference \cite{iclr/SummersD20} empirically observed that $L_2$ regularization on $\gamma$ and $\beta$ often improves performance depending on the architecture.

\begin{table}[h]
	\caption{Preview of our guidelines on $L_2$ regularization.}
	\label{tab:guideline}
	\centering
	\begin{tabular}{l|r|r|r}
		\toprule
		                  & \textbf{TensorFlow}                  & \textbf{PyTorch}                 & \textbf{Our Guidelines}               \\
		\midrule
		$W$               & Applied        \textcolor{blue}{\cmark} & Applied \textcolor{blue}{\cmark} & Applicable \textcolor{blue}{\cmark}  \\
		$\gamma_{last}$   & Rarely applied \textcolor{red}{\xmark}  & Applied \textcolor{blue}{\cmark} & Applicable \textcolor{blue}{\cmark}  \\
		$\gamma_{down}$   & Rarely applied \textcolor{red}{\xmark}  & Applied \textcolor{blue}{\cmark} & Inapplicable \textcolor{red}{\xmark} \\
		$\gamma_{0}$      & Rarely applied \textcolor{red}{\xmark}  & Applied \textcolor{blue}{\cmark} & Inapplicable \textcolor{red}{\xmark} \\
		$\gamma_{others}$ & Rarely applied \textcolor{red}{\xmark}  & Applied \textcolor{blue}{\cmark} & Applicable \textcolor{blue}{\cmark}  \\
		\bottomrule
	\end{tabular}
\end{table}

Because these practices lack theoretical analysis, in this paper, we explore whether it is desirable to apply $L_2$ regularization to the $\gamma$ parameter of BN. First, we claim that $\gamma$ plays a role in controlling variance in residual networks. We show that the variance of the features in the residual block is either accumulated or reset according to the layer arrangement. For better behavior in residual networks, we propose a strategy of making the accumulated variance small and the reset variance large (Section \ref{sec:varianceanalysisinresidualnetworks}). Second, we present an analysis of the effective learning rate from optimization perspective of $\gamma$ (Section \ref{sec:effectivelearningrateanalysis}). Through our theoretical analysis, we present four guidelines for managing $\gamma$ (Table \ref{tab:guideline}).

The validity of the four guidelines is confirmed through several experiments (Section \ref{sec:experiments}). We demonstrate a performance decrease due to incorrect $L_2$ regularization on $\gamma$ and a performance increase due to correct $L_2$ regularization on $\gamma$. We observed this phenomenon in various residual networks and transformers.

\section{Variance analysis in residual networks}
\label{sec:varianceanalysisinresidualnetworks}
In this paper, we target residual networks because they are widely used as a standard architecture and involve many variants. In residual networks, an input image is first passed through the early stage. Then, four stages are applied, and each stage is composed of several residual blocks. Each residual block consists of a skip connection and a residual branch, in which several weight, BN, and ReLU layers are placed in the residual branch.

Our assumption is that the condition when the skip connection dominates over the residual branch, which makes the residual block behave more like an identity mapping, is advantageous for residual network training. In fact, in the original ResNet paper, residual learning was introduced with the intention of modelling identity mappings \cite{gcvpr/HeZRS16}. Reference \cite{gcorr/GoyalDGNWKTJH17} reported that initializing $\gamma=0$ in the last BN causes the residual block to behave like an identity function and eases optimization. Reference \cite{gnips/DeS20} found that a residual block behaves like an identity mapping because the variance of the skip connection is larger than the variance of the residual branch. In their paper, the $\gamma$ parameter of BN was assumed to be 1, but we show here that variance in the residual block varies according to $\gamma$.

We consider the bias parameter $b$ in the weight layers and $\beta$ in the BN layers to be 0. $L_2$ regularization of $b$ or $\beta$ is ignored in this paper. In BN, we assume that the mini-batch size is sufficiently large, and each feature map is normalized to $\mathcal{N}(0, 1)$ and then rescaled using $\gamma$. In other words, the $i$-th element of BN with $\gamma_i$ outputs $B_i(x)$, where $E[B_i(x)] = 0$ and $Var[B_i(x)] = \gamma_i^2$. The term "weight layers" used below indicates convolution layers in ResNet, but will be described by generalizing to fully connected layers. We assume that each element of the weights comes from He initialization $\mathcal{N}(0, 2/fan_{in})$ \cite{giccv/HeZRS15}.

\begin{figure}[t!]
	\centering
    \includegraphics[width=0.74\textwidth]{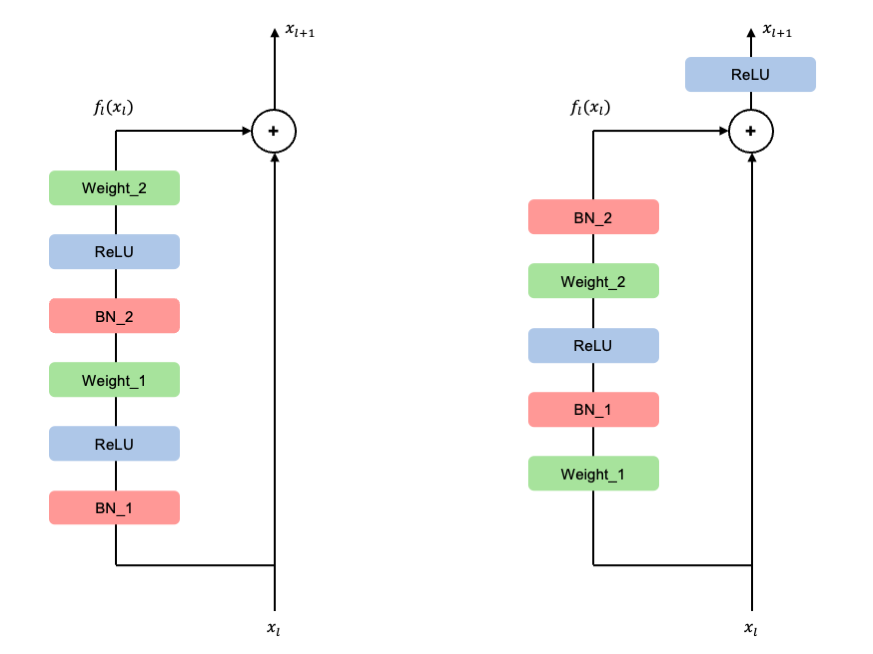}
    \caption{Basic block in (left) PreActResNet and (right) original ResNet (Cases 1 and 2).}
	\label{fig:basic}
\end{figure}

\paragraph{Case 1. Basic block in PreActResNet} \ \\
PreActResNet \cite{geccv/HeZRS16} is a modified version of the original ResNet \cite{gcvpr/HeZRS16}, and is also called ResNetV2. The residual branch of PreActResNet consists of two or three [BN--ReLU--Weight] blocks (Figure \ref{fig:basic}, Left). The $l$-th residual block outputs $x_{l+1}$, which is the sum of the residual branch $f_l(x_l)$ and skip connection $x_l$, i.e., $x_{l+1}=x_l + f_l(x_l)$. Because the covariance between the residual branch and the skip connection is zero \cite{gnips/DeS20}, we have $Var[x_{l+1}]=Var[x_l] + Var[f_l(x_l)]$.

Here, we examine the variance of the residual branch $Var[f_l(x_l)]$. Consider the residual branch with two [BN--ReLU--Weight] blocks. Because of the normalization in the second BN, the $\gamma$ of the first BN and $\left\lVert W_1 \right\rVert$ do not affect the variance of the residual branch. Therefore, the variance of the residual branch is determined by the second [BN--ReLU--Weight] block. By the mean and variance of $W$ from He initialization and BN, the variance of the $i$-th element of the residual branch is
\begin{align}
	Var[f_{l,i}(x_l)] & = \sum_j^{fan_{in}} Var[W_{2,ij,l}] \cdot E[ReLU(B_{2,l,j}(x_l))^2] \\
	                  & = 2 \cdot E[ReLU(B_{2,l,i}(x_l))^2]                                 \\
	                  & = E[B_{2,l,i}(x_l)^2] = Var[B_{2,l,i}(x_l)]                         \\
	                  & = \gamma_{2,l,i}^2. \label{eq:var_res_branch}
\end{align}
The third equality holds because $E[ReLU(X)^2] = \frac{1}{2}E[X^2]$ for normalized $X$ \cite{giccv/HeZRS15}. Thus,
\begin{align}
	Var[x_{l+1,i}] = Var[x_{l,i}] + \gamma_{2,l,i}^2.
\end{align}

Starting from the $s$-th residual block, imagine that the variance of the residual branch is accumulated in a row. The variance of the $l$-th residual block is
\begin{align}
	Var[x_{l,i}] = Var[x_{s,i}] + \sum_{m=s}^{l-1} \gamma_{2,m,i}^2, \label{eq:var_res_block}
\end{align}
where $x_s$ denotes the first input feature map for each stage of ResNet. As such, in the residual block, the variance of the residual branch is accumulated. Here, we want to make the residual block satisfy $Var[x_{l,i}] > Var[f_{l,i}(x_l)]$ so that the skip connection dominates over the residual branch. From Eq. \ref{eq:var_res_branch} and Eq. \ref{eq:var_res_block}, the inequality $Var[x_{s,i}] + \sum_{m=s}^{l-1} \gamma_{2,m,i}^2 > \gamma_{2,l,i}^2$ requires two conditions:
\begin{enumerate}
	\item $\gamma_{2,l,i}^2$ should be small for all $l$;
	\item $Var[x_{s,i}]$ should be large.
\end{enumerate}
The variance at the stage starting point in the second condition is discussed later in Cases 3 and 4. According to the first condition, $\gamma$ in the second BN in all residual blocks should be small to reduce the variance of the residual branch. Similarly, when the residual branch consists of three [BN--ReLU--Weight] blocks, the $\gamma$ in the third BN should be small. Thus, in order for the skip connection to dominate over the residual branch, we should control $\gamma$ in the last BN in the residual branch. To ensure that those called $\gamma_{last}$ are small during training, we recommend applying $L_2$ regularization on $\gamma_{last}$. In summary, we present the following guideline:
\paragraph{Guideline 1.}Because residual block plays the role of variance accumulation, we \textcolor{blue}{should decay} $\gamma_{last}$.

\paragraph{Case 2. Basic block in the original ResNet} \ \\
We check whether Guideline 1 is also applicable to the original ResNet. The original ResNet, also called ResNetV1, constitutes a residual branch with two or three [Weight--BN--ReLU] blocks, but the last ReLU is deployed after addition (Figure \ref{fig:basic}, Right). First, we examine the variance of residual branch $Var[f_l(x_l)]$. Because the residual branch ends with BN, the variance of the residual branch is equal to the variance of the last BN, i.e., $Var[f_{l,i}(x_l)] = \gamma_{last,l,i}^2$.

Contrary to PreActResNet, here, because ReLU is applied after addition, $x_{l+1} = ReLU(x_l + f_l(x_l))$. This ReLU after addition makes a different variance accumulation. Ignoring $E[x_{l,i}]$, we have
\begin{align}
	Var[x_{l+1,i}] & = \frac{1}{2} E[(x_{l,i} + f_{l,i}(x_l))^2]                                  \\
	               & = \frac{1}{2} (E[x_{l,i}^2] + E[f_{l,i}(x_l)^2])                             \\
	               & = \frac{1}{2} (Var[x_{l,i}] + Var[f_{l,i}(x_l)])                             \\
	               & = \frac{1}{2} (Var[x_{l,i}] + \gamma_{last,l,i}^2). \label{eq:half_variance}
\end{align}

Thus, the added variance is halved. Here, we introduce the following substitutions: $a_l = 2^l Var[x_{l,i}]$ and $b_l = 2^l \gamma_{last,l,i}^2$. Rewriting Eq. \ref{eq:half_variance}, we have $a_{l+1} = a_l + b_l$, and thus $a_l = a_s + \sum_{m=s}^{l-1} b_m$. Therefore, we obtain
\begin{align}
    Var[x_{l,i}] & = \frac{1}{2^{l-s}} Var[x_{s,i}] + \sum_{m=s}^{l-1} \frac{1}{2^{l-m}} \gamma_{last,m,i}^2. \label{eq:half_variance_lth}
\end{align}

As such, the variance of the residual branch is accumulated here as well, but because the variance of the last ReLU is halved, the variance of the previous block decays. We call this \emph{half variance accumulation}. For the skip connection to dominate, the condition $Var[x_{l,i}] > Var[f_{l,i}(x_l)]$ requires that 1) $\gamma_{last,l,i}^2$ should be small and 2) $Var[x_{s,i}]$ should be large. Therefore, it is desirable to decay $\gamma_{last}$, and thus, Guideline 1 is valid for the original ResNet as well.

\begin{figure}[t!]
	\centering
    \includegraphics[width=0.74\textwidth]{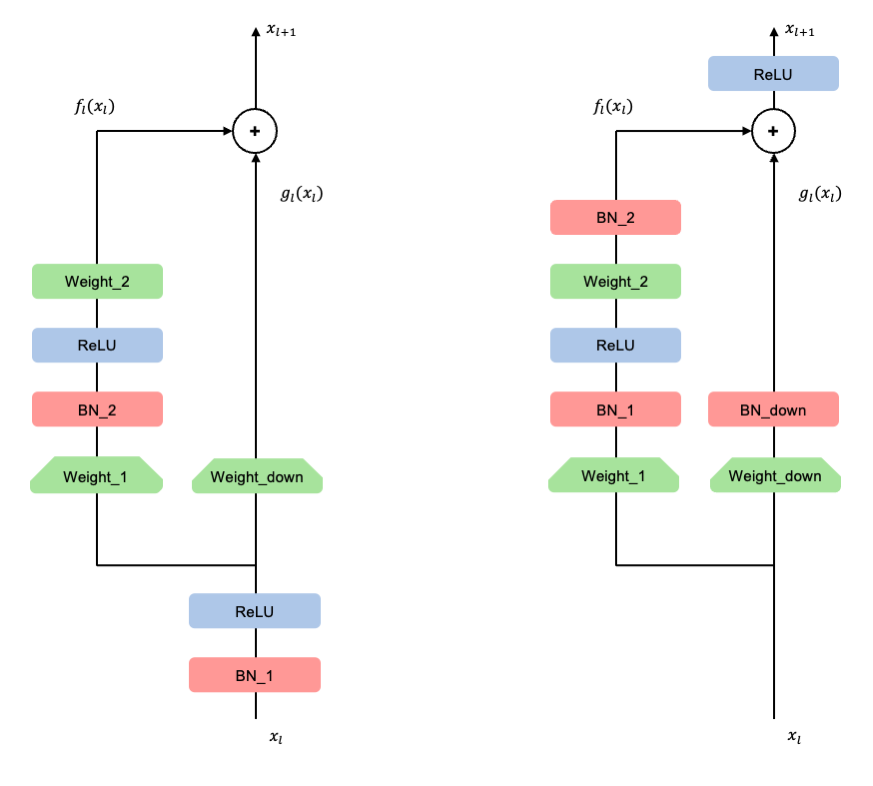}
	\caption{Downsampling block in (left) PreActResNet and (right) original ResNet (Case 3).}
	\label{fig:downsample}
\end{figure}

\paragraph{Case 3. Downsampling block in ResNet} \ \\
Convolutional neural networks downsample high-dimensional image features to low-dimensional ones using pooling layers \cite{DBLP:conf/nips/KrizhevskySH12}. For ResNet, downsampling is performed using strided convolution \cite{DBLP:journals/corr/SpringenbergDBR14} as well as pooling. For the original ResNet, at the end of each stage, a downsampling block (Figure \ref{fig:downsample}, Right) is applied, where the strided convolution is used in both the first weight layers in the residual branch and the skip path. Here, the skip path of the downsampling block is composed of strided convolution and BN rather than identity. We denote the skip path output as $g_l(x_l)=B_{down,l}(W_{down,l}x_l)$. Because each branch ends with BN, the variance of each branch is determined by the last BN as follows:
\begin{align}
	Var[g_{l,i}(x_l)] & = \gamma_{down,l,i}^2, \\
	Var[f_{l,i}(x_l)] & = \gamma_{last,l,i}^2.
\end{align}

From $x_{l+1} = ReLU(g_l(x_l) + f_l(x_l))$, the variance after the downsampling block is
\begin{align}
	Var[x_{l+1,i}] & = \frac{1}{2} (Var[g_{l,i}(x_l)] + Var[f_{l,i}(x_l)])      \\
	               & = \frac{1}{2} (\gamma_{down,l,i}^2 + \gamma_{last,l,i}^2).
\end{align}
Note that the output variance of the downsampling block is not affected by the input variance and is newly determined by $\gamma_{down}$ and $\gamma_{last}$. We call this \emph{variance reset}.

Again, it is desirable to decay $\gamma_{last}$ so that the residual branch does not dominate. However, it is not desirable to decay $\gamma_{down}$. There are two reasons for this. First, in order for the skip path in the downsampling block to dominate, we should decay $\gamma_{last}$ and should not decay $\gamma_{down}$. Second, it is desirable to have a large reset variance. The output variance of the downsampling block becomes the input variance of the successive $(l+1)$-th residual block, which is the variance at the stage starting point. In Cases 1 and 2, we confirmed that a large $Var[x_{s,i}]$ is favorable. Thus, a large output variance of the downsampling block is desirable. In other words, if we decay both $\gamma_{down}$ and $\gamma_{last}$, the reset variance decreases, which is undesirable. Therefore, we claim that $\gamma_{down}$ should not be subjected to $L_2$ regularization to ensure that the reset variance appears dominant afterward. In this regard, we present the following guideline:
\paragraph{Guideline 2.}Because the downsampling block plays the role of variance reset, we \textcolor{red}{should not decay} $\gamma_{down}$.

\begin{wrapfigure}{r}{0.26\textwidth}
    \centering
    \includegraphics[width=0.25\textwidth]{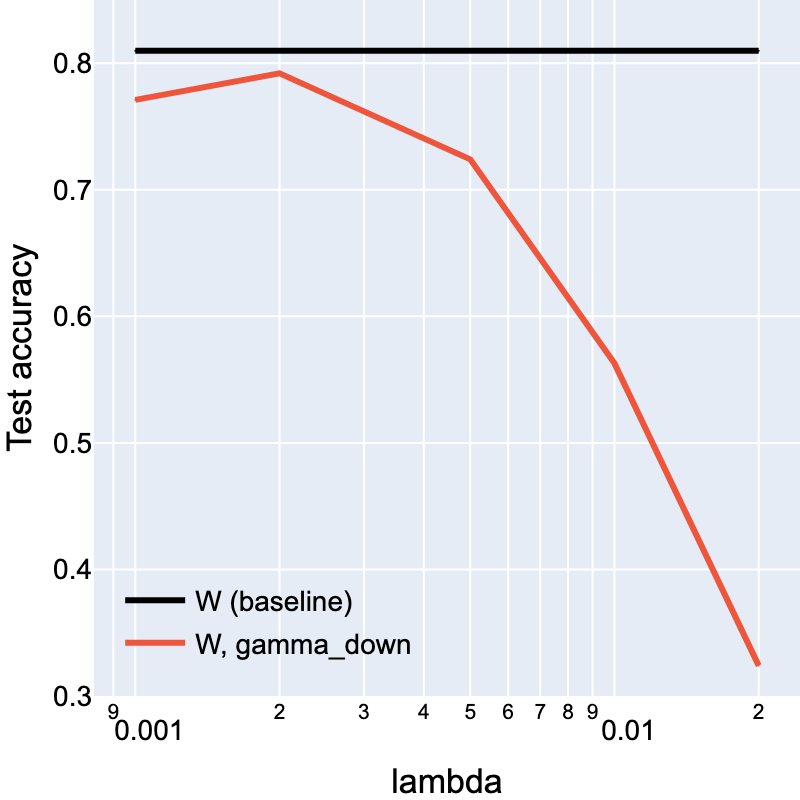}
    \caption{Performance degradation caused by $L_2$ regularization on $\gamma_{down}$ with various $\lambda$ for ResNet-152 on the PET dataset.}
    \label{fig:lambda}
\end{wrapfigure}
Now, we check whether Guideline 2 is also applicable to the downsampling block of PreActResNet. In the downsampling block of PreActResNet, before starting both the residual branch and skip path, block input $x_l$ is subjected to the first BN and ReLU (Figure \ref{fig:downsample}, Left). The first BN does not affect the variance of the residual branch but does affect the variance of the skip path. For skip path $g_l(x_l) = W_{down,l} ReLU(B_{1,l}(x_l))$, we have $Var[g_{l,i}(x_l)] = 2 \cdot E[ReLU(B_{1,l,i}(x_l))^2] = \gamma_{1,l,i}^2$. For the residual branch, similar to Case 1, we have $Var[f_{l,i}(x_l)] = \gamma_{2,l,i}^2$. Thus, the variance after residual block is
\begin{align}
	Var[x_{l+1,i}] & = Var[g_{l,i}(x_l)] + Var[f_{l,i}(x_l)] \\
	               & = \gamma_{1,l,i}^2 + \gamma_{2,l,i}^2.
\end{align}

Thus, the downsampling block of PreActResNet plays the role of variance reset as well. To ensure that the reset variance is dominant afterward, we should not decay $\gamma_1$ at the first BN. For convenience, we also call this parameter $\gamma_{down}$.

\begin{figure}[t!]
	\centering
    \includegraphics[width=0.30\textwidth]{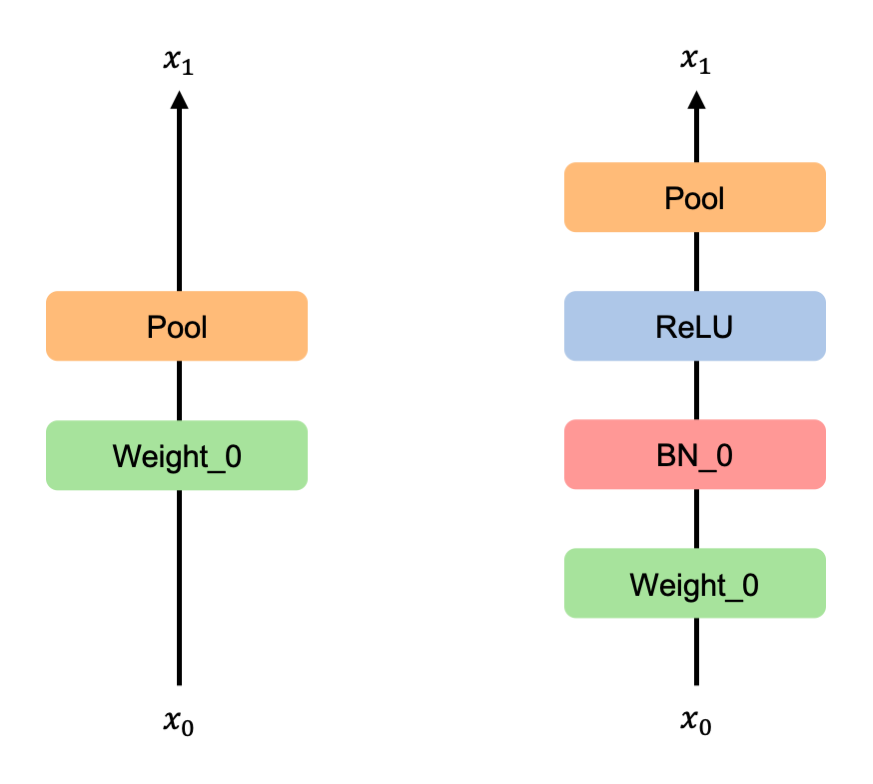}
    \caption{Early stage in (left) PreActResNet and (right) original ResNet (Case 4).}
	\label{fig:early}
\end{figure}

\paragraph{Case 4. Early stage in ResNet} \ \\
The input variance of the first residual block is determined by the early stage of ResNet before the residual block starts. The early stage of ResNet, which is also called the stem, consists of convolution, BN, ReLU, and pooling layers without a skip connection in the original ResNet (Figure \ref{fig:early}, Right). Ignoring pooling, from the output of early stage $x_1 = ReLU(B_0(W_0 x_0))$, we have
\begin{align}
	Var[x_{1,i}] & = \frac{1}{2} E[B_{0,i}(W_0 x_0))^2] \\
	             & = \frac{1}{2} \gamma_{0,i}^2.
\end{align}
Therefore, $\gamma_0$ \emph{sets} the output variance of the early stage. The output of the early stage becomes the input of the residual block of the first stage. Because a large variance at the stage starting point is favorable, $\gamma_0$ should be large. In this regard, we present the following guideline:
\paragraph{Guideline 3.}Because the early stage plays the role of variance set, we \textcolor{red}{should not decay} $\gamma_0$.

In case of PreActResNet, BN is not applied in the early stage (Figure \ref{fig:early}, Left). Thus, we do not have to consider $\gamma_0$.

Finally, in the residual branch, additional $\gamma$ parameters exist in the BN before the last BN for both PreActResNet and the original ResNet. We call those parameters $\gamma_{others}$. Because they do not affect the variance of the residual branches, our variance analysis cannot explain their role. The $\gamma_{others}$ parameters are discussed again in Section \ref{sec:effectivelearningrateanalysis}; here, we simply present the conclusion:
\paragraph{Guideline 4.}Though other BNs do not determine the variance, $L_2$ regularization on $\gamma_{others}$ helps optimization by improving the effective learning rate. Thus, we \textcolor{blue}{should decay} $\gamma_{others}$.

\begin{wrapfigure}{r}{0.26\textwidth}
	\centering
    \includegraphics[width=0.25\textwidth]{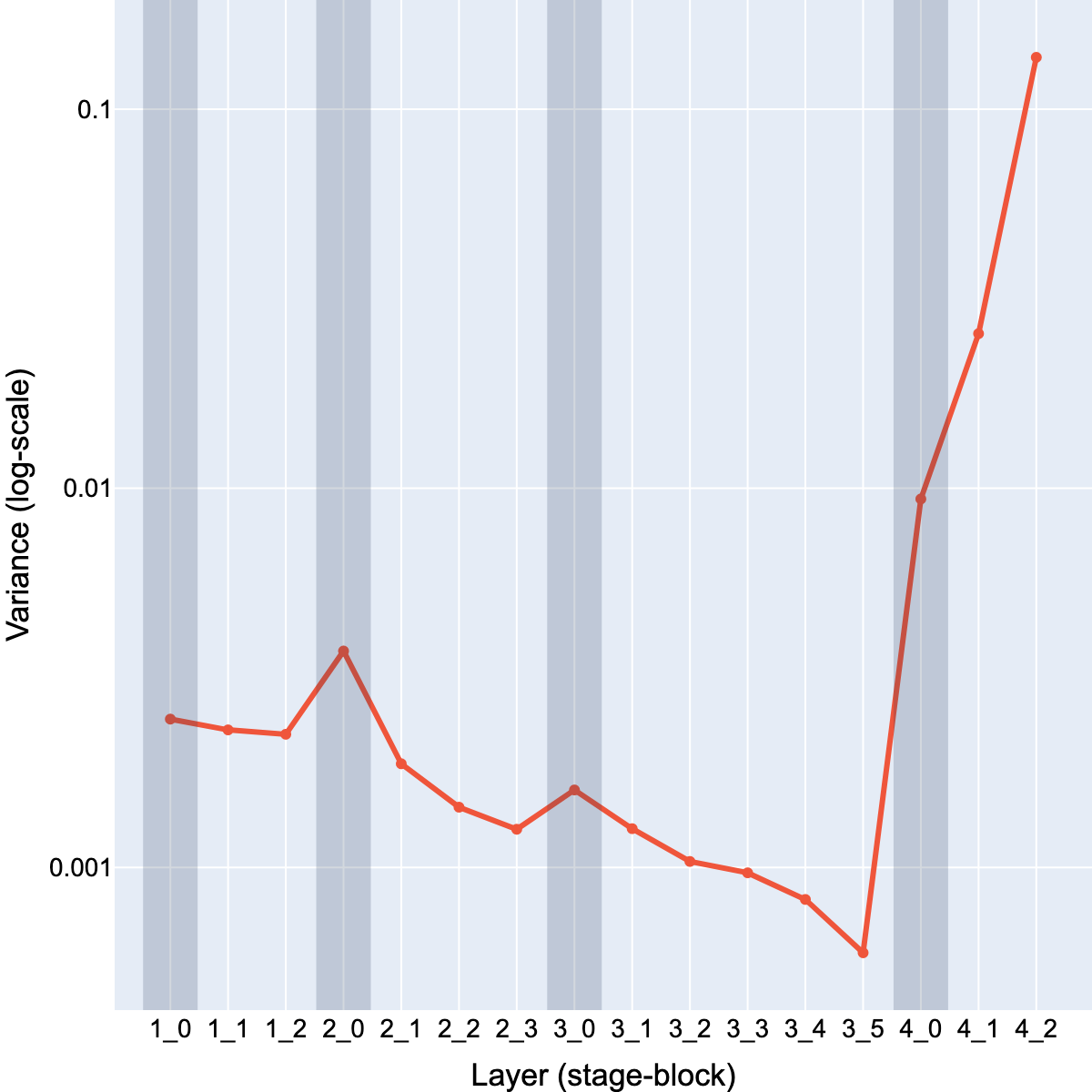}
    \caption{Variance of each feature map of ResNet-50 pre-trained on ImageNet.}
    \label{fig:variance}
\end{wrapfigure}

Further, using ResNet-50, we measured the variance of each feature map (Figure \ref{fig:variance}). First, the variance is (re)set at layer 1-0, and then the variance decreases at layer 1-1, which is expected from the half variance accumulation (Eqs. \ref{eq:half_variance} and \ref{eq:half_variance_lth}). In layer 2-0, where the subsequent stage begins, the variance is reset to a new larger value and decreases again from layer 2-1 to layer 2-3. As such, both the variance reset to a larger value and half variance accumulation with decreasing value appear in each residual block, which agrees with our claim.

\paragraph{Case 5. Transformer block} \ \\
The transformer is attracting attention for its high performance in various tasks, including computer vision \cite{giclr/DosovitskiyB0WZ21} and natural language processing \cite{gnips/VaswaniSPUJGKP17,gnaacl/DevlinCLT19}. The transformer uses a series of transformer blocks, which consists of residual branches with self-attention and multilayer perceptron blocks. In contrast to ResNet, the transformer uses LayerNorm (LN) \cite{gcorr/BaKH16}. However, LN also rescales normalized features using $\gamma$ and $\beta$. Therefore, our variance analysis can explain how the $\gamma$ parameter of LN affects the variance within the transformer block. See the Appendix for the mathematical details regarding the roles of $\gamma_1$ and $\gamma_2$ in the first and second LNs. Here, we simply present the following guideline for the transformer block:
\paragraph{Guideline 1T.}Because a transformer block accumulates the variance, we \textcolor{blue}{should decay} $\gamma_1$ and $\gamma_2$.

Note that the transformer has no downsampling block (and hence there is no Guideline 2T). Further, some transformers such as bidirectional encoder representations from transformers (BERT) \cite{gnaacl/DevlinCLT19} have LN in the early stage, whose $\gamma_0$ should not be subjected to the $L_2$ regularization (Guideline 3T). Finally, some transformers \cite{gnips/VaswaniSPUJGKP17} have another LN between the final transformer block and the head layer. The role of the $\gamma_{other}$ parameters of LN cannot be explained by our variance analysis, but in the same context as Guideline 4, $L_2$ regularization on $\gamma_{other}$ helps optimization (Guideline 4T).

\section{Effective learning rate analysis}
\label{sec:effectivelearningrateanalysis}
In this section, we explore why $\gamma_{others}$ parameters should be decayed even though their scale does not affect the BN output. Reference \cite{gcorr/Laarhoven17b,gnips/HofferBGS18,giclr/ZhangWXG19} argued that in a neural network with BN, the weight scale does not affect the BN output. They introduced the \emph{effective learning rate} to claim that a smaller weight scale is advantageous for optimization, which can be achieved through $L_2$ regularization on weights. In our paper, we note that $\gamma$ is a trainable parameter in a neural network similar to weights. Extending their claim, we investigate the effective learning rate for the optimization of $\gamma$.

BN is composed of a normalization step $N(\cdot)$ and linear scaling step $L(\cdot)$. From a series of [Weight--BN--ReLU] blocks, we decompose BN into [N--L] and investigate the intermediate [L--ReLU--Weight--N] block. In the $l$-th intermediate block, weight layer holds $W_l$ and $L$ step holds $\Gamma_l$, where $\Gamma_l$ is a diagonal matrix with $(\Gamma_l)_{i,i}=\gamma_i$ and $(\Gamma_l)_{i,j}=0$ for $i \neq j$. From the input feature map $x_l$, the intermediate block outputs $x_{l+1} = y(x_l ; W_l, \Gamma_l) = N[W_l ReLU(\Gamma_l x_l)]$.

Here, we decompose the diagonal matrix into its norm and direction, i.e., $\Gamma_l = \left\lVert \Gamma_l \right\rVert \hat{\Gamma}_l$. Then,
\begin{align}
	y(x_l ; W_l, \Gamma_l) & = N[W_l ReLU(\left\lVert \Gamma_l \right\rVert \hat{\Gamma}_l x_l)] = N[W_l \left\lVert \Gamma_l \right\rVert ReLU(\hat{\Gamma}_l x_l)] \\
	                       & = N[W_l ReLU(\hat{\Gamma}_l x_l)].
\end{align}

Thus, the scale of $\gamma$ in BN does not affect the intermediate block output. If we investigate the gradient of $y(x_l ; W_l, \Gamma_l)$ with respect to $\Gamma_l$, we obtain
\begin{align}
	\nabla_{\Gamma_l} y(x_l ; W_l, \Gamma_l) = \frac{1}{\left\lVert \Gamma_l \right\rVert} \nabla_{\hat{\Gamma}_l} y(x_l ; W_l, \hat{\Gamma}_l). \label{eq:grad_scale}
\end{align}

This scale-variant gradient property was used in \cite{gcorr/Laarhoven17b,gnips/HofferBGS18} to claim that $L_2$ regularization on weights helps optimization even in neural networks with BN. Thus, effective learning rate analysis is expected to be applicable to $\gamma$ in BN. From Eq. \ref{eq:grad_scale}, we have $\nabla_{\Gamma_l} L(\Gamma_l) = \frac{1}{\left\lVert \Gamma_l \right\rVert} \nabla_{\hat{\Gamma}_l} L(\hat{\Gamma}_l)$ for loss function $L$. First, gradient descent with respect to $\Gamma_l$ is in the form of
\begin{align}
	\Gamma_{l,t+1} = \Gamma_{l,t} - \eta \nabla_{\Gamma_{l,t}} L(\Gamma_{l,t}).
\end{align}

If we investigate gradient descent with respect to $\hat{\Gamma}_l$, we have,
\begin{align}
    \hat{\Gamma}_{l,t+1} = \hat{\Gamma}_{l,t} - \left(\eta \left\lVert \Gamma_{l,t} \right\rVert^{-2} \right) \nabla_{\hat{\Gamma}_{l,t}} L(\hat{\Gamma}_{l,t}). \label{eq:gamma_dir_sgd}
\end{align}

\begin{wrapfigure}{r}{0.26\textwidth}
	\centering
    \includegraphics[width=0.25\textwidth]{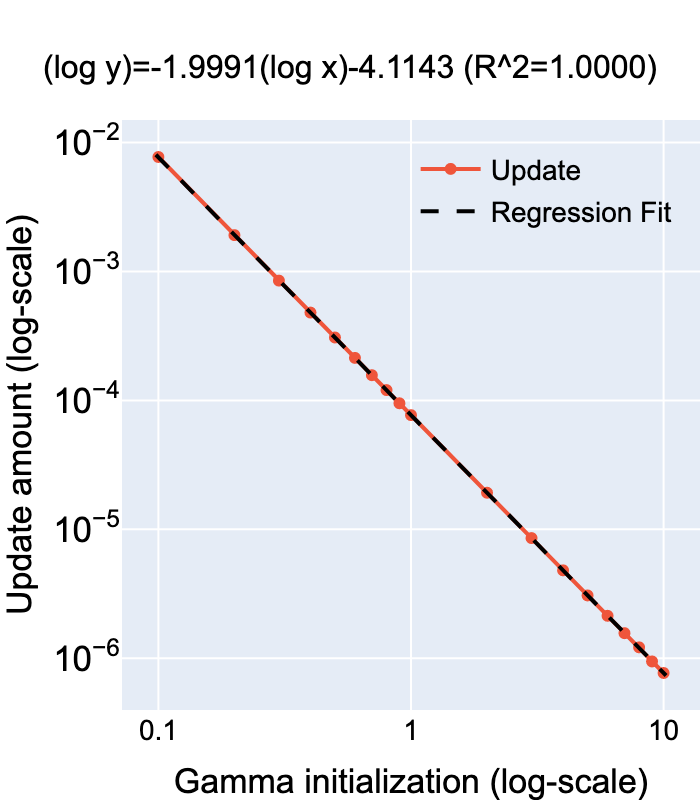}
    \caption{Norm of the first update by the initial value of $\gamma$.}
    \label{fig:update}
\end{wrapfigure}

Here, $\eta \left\lVert \Gamma_{l,t} \right\rVert^{-2}$ is called the effective learning rate \cite{gcorr/Laarhoven17b,gnips/HofferBGS18}. If the scale of $\gamma$ in BN decreases, the effective learning rate increases, which prevents the optimization from saturating due to the large weight norm. Although we followed the derivation in \cite{gcorr/Laarhoven17b}, the derivation in \cite{gnips/HofferBGS18} shows similar results. Thus, although $L_2$ regularization on $\gamma_{others}$ does not affect the variance, it improves the effective learning rate and optimization. In this regard, we advocate applying $L_2$ regularization to $\gamma_{others}$.

Now, we empirically validate Eq. \ref{eq:gamma_dir_sgd}. Using ResNet-18, we varied the initial value of $\gamma$ and measured the norm of the first update, $\lVert \hat{\Gamma}_{l,1} -\hat{\Gamma}_{l,0} \rVert$ (Figure \ref{fig:update}). From their log values, a regression fit yielded a coefficient near $-2$, which agrees with $\lVert \hat{\Gamma}_{l,t+1} -\hat{\Gamma}_{l,t} \rVert \propto \lVert \Gamma_{l,t} \rVert^{-2}$.

Note that this effective learning rate analysis is applicable to the $\gamma$ parameters of all four categories. Although applying $L_2$ regularization on $\gamma_{down}$ and $\gamma_{0}$ is harmful to variance control, it could be advantageous for improving optimization with respect to the effective learning rate.

\section{Experiments}
\label{sec:experiments}

\subsection{Image classification tasks}
We tested whether Guidelines 1--4 are valid in practical tasks. First, in the training task of ResNet and its variants \cite{gcvpr/HeZRS16,geccv/HeZRS16,gbmvc/ZagoruykoK16,gcvpr/XieGDTH17}, we measured the performance of applying $L_2$ regularization to the weights and $\gamma$ of each group with $\frac{1}{2} \lambda (\left\lVert W \right\rVert^2_2 + \left\lVert \gamma \right\rVert^2_2)$. For all the experiments, we report the average results from three experiments. For hyperparameters and training details, refer to the Appendix.

\begin{table}[h!]
	\caption{Test accuracy (\%) for the PET dataset.}
	\label{tab:pet}
	\centering
	\begin{tabular}{l|r|r|r|r|r}
		\toprule
		$\pmb{L_2}$ \textbf{Regularization} & $\pmb{W}$ & $\pmb{W,\gamma_{last}}$     & $\pmb{W, \gamma_{down}}$   & $\pmb{W, \gamma_0}$         & $\pmb{W, \gamma_{others}}$  \\
		\midrule
		\midrule
		ResNet-18                           & 83.5733   & 85.4100                     & 82.5800                    & 82.8800                     & 84.6000                     \\
		                                    &           & \textcolor{blue}{(+1.8367)} & \textcolor{red}{(-0.9933)} & \textcolor{red}{(-0.6933)}  & \textcolor{blue}{(+1.0267)} \\
		\midrule
		ResNet-50                           & 82.9433   & 85.6500                     & 80.0833                    & 82.4000                     & 83.5133                     \\
		                                    &           & \textcolor{blue}{(+2.7067)} & \textcolor{red}{(-2.8600)} & \textcolor{red}{(-0.5433)}  & \textcolor{blue}{(+0.5700)} \\
		\midrule
		ResNet-101                          & 82.0100   & 84.4200                     & 78.9400                    & 81.3767                     & 83.1200                     \\
		                                    &           & \textcolor{blue}{(+2.4100)} & \textcolor{red}{(-3.0700)} & \textcolor{red}{(-0.6333)}  & \textcolor{blue}{(+1.1100)} \\
		\midrule
		ResNet-152                          & 80.9867   & 85.7400                     & 79.2100                    & 82.3700                     & 82.9100                     \\
		                                    &           & \textcolor{blue}{(+4.7533)} & \textcolor{red}{(-1.7767)} & \textcolor{blue}{(+1.3833)} & \textcolor{blue}{(+1.9233)} \\
		\midrule
		Wide ResNet-50-2                    & 82.9433   & 85.6500                     & 80.0833                    & 82.4000                     & 83.5133                     \\
		                                    &           & \textcolor{blue}{(+2.7067)} & \textcolor{red}{(-2.8600)} & \textcolor{red}{(-0.5433)}  & \textcolor{blue}{(+0.5700)} \\
		\midrule
		Wide ResNet-101-2                   & 82.4600   & 85.1400                     & 81.2000                    & 83.4533                     & 84.6300                     \\
		                                    &           & \textcolor{blue}{(+2.6800)} & \textcolor{red}{(-1.2600)} & \textcolor{blue}{(+0.9933)} & \textcolor{blue}{(+2.1700)} \\
		\midrule
		ResNeXt-50-32x4d                    & 84.4167   & 85.6200                     & 84.1133                    & 84.6000                     & 85.7400                     \\
		                                    &           & \textcolor{blue}{(+1.2033)} & \textcolor{red}{(-0.3033)} & \textcolor{blue}{(+0.1833)} & \textcolor{blue}{(+1.3233)} \\
		\midrule
		ResNeXt-101-32x8d                   & 84.0233   & 85.4100                     & 83.6067                    & 82.9700                     & 85.2900                     \\
		                                    &           & \textcolor{blue}{(+1.3867)} & \textcolor{red}{(-0.4167)} & \textcolor{red}{(-1.0533)}  & \textcolor{blue}{(+1.2667)} \\
        \midrule
		\midrule
        Our Guidelines                      & \textcolor{blue}{\cmark} & \textcolor{blue}{\cmark}    & \textcolor{red}{\xmark}    & \textcolor{red}{\xmark}     & \textcolor{blue}{\cmark}    \\
		\bottomrule
	\end{tabular}
\end{table}

Table \ref{tab:pet} shows image classification accuracy on the Oxford-IIIT PET dataset \cite{gcvpr/ParkhiVZJ12} for different $L_2$ regularization setups. First, when the $\gamma_{last}$ parameters were subjected to $L_2$ regularization, the test accuracy increased approximately 1\%--4\% compared to the case where $L_2$ regularization was not applied to them. This is consistent with Guideline 1, which states that we should decay the $\gamma_{last}$. However, when $L_2$ regularization was applied to $\gamma_{down}$, the test accuracy decreased approximately 1\%--3\%. This is consistent with Guideline 2, which states that $L_2$ regularization for $\gamma_{down}$ is undesirable. Similarly, for $\gamma_0$ and $\gamma_{others}$, the experimental results agree with Guidelines 3 and 4.

\paragraph{Limitations} However, for $\gamma_0$, an increase in performance was observed in some models. When the effect of improving the effective learning rate is more dominant than the variance control, it can result in performance improvement. Because the first stage is relatively shallow compared to other stages, the variance from $\gamma_0$ would have a minor effect, and improvement of the effective learning rate may become dominant. The coexistence of variance control and improved effective learning rate requires further research. See the Appendix for more results on other datasets, including NABirds, Food-101, and CIFAR-10.

\subsection{Natural language processing tasks}
We also tested whether our proposed guidelines are valid for transformers on natural language processing tasks. First, we visited a machine translation task using a transformer. The BLEU score \cite{acl/PapineniRWZ02} was measured for the German to English translation task using the IWSLT-14 dataset \cite{cettolo2014report} (Table \ref{tab:iwslt}). We measured the performance before and after applying $L_2$ regularization to $\gamma_1$ and $\gamma_2$. For both cases, an increase in the BLEU score was observed. This result is consistent with Guideline 1T, which states that small $\gamma$ is desirable for transformer blocks because they accumulate variance.

\begin{table}[h]
	\caption{BLEU scores (\%) for the IWSLT-14 dataset.}
	\label{tab:iwslt}
	\centering
	\begin{tabular}{l|r|r|r}
		\toprule
		$\pmb{L_2}$ \textbf{Regularization} & $\pmb{W}$ & $\pmb{W,\gamma_1}$          & $\pmb{W,\gamma_2}$          \\
		\midrule
		Transformer                         & 34.7894   & 34.9063                     & 35.1385                     \\
		                                    &           & \textcolor{blue}{(+0.1170)} & \textcolor{blue}{(+0.3491)} \\
		\bottomrule
	\end{tabular}
\end{table}

Our claim was also verified in the text classification task using BERT for the SST-2 task from GLUE \cite{giclr/WangSMHLB19}. Following Guideline 1T improved test accuracy here as well (Table \ref{tab:sst2}).

\begin{table}[h]
	\caption{Test accuracy (\%) for the GLUE SST-2 task.}
	\label{tab:sst2}
	\centering
	\begin{tabular}{l|r|r|r}
		\toprule
		$\pmb{L_2}$ \textbf{Regularization} & $\pmb{W}$ & $\pmb{W,\gamma_1,\gamma_2}$ & \textbf{Difference}       \\
		\midrule
		BERT                                & 91.8807   & 92.0872                     & \textcolor{blue}{+0.2064} \\
		\bottomrule
	\end{tabular}
\end{table}

\section{Conclusion}
\label{sec:conclusion}
In this paper, we studied $L_2$ regularization for the $\gamma$ parameters of BN. We theoretically showed that this $\gamma$ should be controlled so that the residual blocks behave similarly to identity mapping. Specifically, we discussed the cases in which it is desirable to decay the $\gamma$ parameters of BN and the cases in which it is not, presenting four guidelines for their management. The proposed guidelines were verified in experiments on several variants of residual networks and transformer models. We confirmed that the performance increases when the guidelines are followed and decreases when the guidelines are violated.

\newpage

\bibliography{gamma_5}
\bibliographystyle{plain}

\newpage

\section*{Checklist}


\begin{enumerate}

	\item For all authors...
	      \begin{enumerate}
		      \item Do the main claims made in the abstract and introduction accurately reflect the paper's contributions and scope?
		            \answerYes{}
		      \item Did you describe the limitations of your work?
		            \answerYes{We discussed limitations and future research in the third paragraph of Section \ref{sec:experiments}}
		      \item Did you discuss any potential negative societal impacts of your work?
		            \answerNA{}
		      \item Have you read the ethics review guidelines and ensured that your paper conforms to them?
		            \answerYes{}
	      \end{enumerate}

	\item If you are including theoretical results...
	      \begin{enumerate}
		      \item Did you state the full set of assumptions of all theoretical results?
		            \answerYes{See Section \ref{sec:varianceanalysisinresidualnetworks}.}
		      \item Did you include complete proofs of all theoretical results?
		            \answerYes{See Section \ref{sec:varianceanalysisinresidualnetworks} and the Appendix.}
	      \end{enumerate}

	\item If you ran experiments...
	      \begin{enumerate}
		      \item Did you include the code, data, and instructions needed to reproduce the main experimental results (either in the supplemental material or as a URL)?
		            \answerYes{See the supplemental material.}
		      \item Did you specify all the training details (e.g., data splits, hyperparameters, how they were chosen)?
		            \answerYes{See the Appendix.}
		      \item Did you report error bars (e.g., with respect to the random seed after running experiments multiple times)?
		            \answerNo{Though we did not use error bars, we report the average from three results for all experiments.}
		      \item Did you include the total amount of compute and the type of resources used (e.g., type of GPUs, internal cluster, or cloud provider)?
		            \answerYes{See the Appendix.}
	      \end{enumerate}

	\item If you are using existing assets (e.g., code, data, models) or curating/releasing new assets...
	      \begin{enumerate}
		      \item If your work uses existing assets, did you cite the creators?
		            \answerYes{The data and models are properly cited.}
		      \item Did you mention the license of the assets?
		            \answerNo{Though we did not mention the licenses, we used existing assets, which are available online and do not have any license issues.}
		      \item Did you include any new assets either in the supplemental material or as a URL?
		            \answerYes{See the supplemental material.}
		      \item Did you discuss whether and how consent was obtained from people whose data you're using/curating?
		            \answerNo{We used existing assets which are available online and do not have any issues, such as the PyTorch library and open datasets.}
		      \item Did you discuss whether the data you are using/curating contains personally identifiable information or offensive content?
		            \answerNo{Though we did not discuss such issues, the assets we used do not have any critical issues such as information about a specific person.}
	      \end{enumerate}

	\item If you used crowdsourcing or conducted research with human subjects...
	      \begin{enumerate}
		      \item Did you include the full text of instructions given to participants and screenshots, if applicable?
		            \answerNA{}
		      \item Did you describe any potential participant risks, with links to Institutional Review Board (IRB) approvals, if applicable?
		            \answerNA{}
		      \item Did you include the estimated hourly wage paid to participants and the total amount spent on participant compensation?
		            \answerNA{}
	      \end{enumerate}

\end{enumerate}

\end{document}